\documentclass[11pt,a4paper]{article}
\usepackage[hyperref]{emnlp2020}
\usepackage{times}
\usepackage{latexsym}

\usepackage{microtype}

\aclfinalcopy % Uncomment this line for the final submission
\usepackage[utf8]{inputenc} % allow utf-8 input
\usepackage[T1]{fontenc}    % use 8-bit T1 fonts
\usepackage{hyperref}       % hyperlinks
\usepackage{url}            % simple URL typesetting
\usepackage{booktabs}       % professional-quality tables
\usepackage{amsfonts}       % blackboard math symbols
\usepackage{nicefrac}       % compact symbols for 1/2, etc.
\usepackage{microtype}      % microtypography
\usepackage{amsmath}
\usepackage{graphicx}

\usepackage{xcolor}
\usepackage{wrapfig}
\usepackage{comment}
\usepackage{tabu}

\usepackage{caption}
\usepackage{subfigure}

\usepackage{multicol}

\usepackage{longtable}
\usepackage{makecell}
\usepackage{amssymb}
\usepackage{arydshln }
\usepackage{paralist}

\usepackage{bm}
\newcommand{\vect}[1]{\boldsymbol{\mathbf{#1}}}
\usepackage{amsmath,amsfonts,bm}

\def\eqref#1{equation~\ref{#1}}
\def\1{\bm{1}}

\def\rvh{{\mathbf{h}}}
\def\rvu{{\mathbf{i}}}

\def\rvu{{\mathbf{u}}}

\def\rvw{{\mathbf{w}}}

\def\rvz{{\mathbf{z}}}

\DeclareMathAlphabet{\mathsfit}{\encodingdefault}{\sfdefault}{m}{sl}
\SetMathAlphabet{\mathsfit}{bold}{\encodingdefault}{\sfdefault}{bx}{n}

\newcommand*{\mathcolor}{}
\def\mathcolor#1#{\mathcoloraux{#1}}
\newcommand*{\mathcoloraux}[3]{%
  \protect\leavevmode
  \begingroup
    \color#1{#2}#3%
  \endgroup
}

\definecolor{darkblue}{rgb}{0.0, 0.0, 0.55}
\hypersetup{
    colorlinks=true,
    citecolor=darkblue
}

\definecolor{Orange}{RGB}{255, 89, 66}

\title{On the Sentence Embeddings from Pre-trained Language Models}
\author{Bohan Li$^{\dagger,\ddagger}$\thanks{\ \ The work was done when BL was an intern at ByteDance.} , 
Hao Zhou$^\dagger$, 
Junxian He$^\ddagger$, 
Mingxuan Wang$^\dagger$, 
Yiming Yang$^\ddagger$, 
Lei Li$^\dagger$ \\
  $^\dagger$ByteDance AI Lab \\
  $^\ddagger$Language Technologies Institute, Carnegie Mellon University \\
  {\texttt{\{zhouhao.nlp,wangmingxuan.89,lileilab\}@bytedance.com}} \\
  {\texttt{\{bohanl1,junxianh,yiming\}@cs.cmu.edu}}
  }

\date{}

\begin{document}
\maketitle

\begin{abstract}
Pre-trained contextual representations like BERT have achieved great success in natural language processing. 
However, the sentence embeddings from the pre-trained language models without fine-tuning have been found to poorly capture semantic meaning of sentences. 
In this paper, we argue that the semantic information in the BERT embeddings is not fully exploited. 
We first reveal the theoretical connection between the masked language model pre-training objective and the semantic similarity task theoretically, and then analyze the BERT sentence embeddings empirically. We find that BERT always induces a non-smooth anisotropic semantic space of sentences, which harms its performance of semantic similarity. To address this issue, we propose to transform the anisotropic sentence embedding distribution to a smooth and isotropic Gaussian distribution through normalizing flows that are learned with an unsupervised objective. Experimental results show that our proposed BERT-flow method obtains significant performance gains over the state-of-the-art sentence embeddings on a variety of semantic textual similarity tasks.
The code is available at \url{https://github.com/bohanli/BERT-flow}.

\end{abstract}

\section{Introduction}
\label{sec:intro}
Recently, pre-trained language models and its variants~\citep{radford2019language,devlin2018bert,yang2019xlnet,liu2019roberta} like BERT~\citep{devlin2018bert} have been widely used as representations of natural language. Despite their great success on many NLP tasks through fine-tuning, the sentence embeddings from BERT without fine-tuning are significantly inferior in terms of semantic textual similarity~\citep{reimers2019sentence} -- for example, they even underperform the GloVe~\citep{pennington2014glove} embeddings which are not contextualized and trained with a much simpler model. Such issues hinder applying BERT sentence embeddings directly to many real-world scenarios where collecting labeled data is highly-costing or even intractable.

% and tree-based representations~\citep{shi2018tree}
% such as clustering, information retrieval, or low-resource languages .

In this paper, we aim to answer two major questions: (1) why do the BERT-induced sentence embeddings perform poorly to retrieve semantically similar sentences? Do they carry too little semantic information, or just because the semantic meanings in these embeddings are not exploited properly? (2) If the BERT embeddings capture enough semantic information that is hard to be directly utilized, how can we make it easier without external supervision?
% However, we usually need to fine-tune BERT in the downstream task with a number of labeled examples practically to achieve the cutting-edge performance in downstream tasks~\citep{devlin2018bert,liu2019multi,dai2019deeper,reddy2019coqa}. Collecting labeled data may be highly-costing or even intractable in many real-world scenarios.

% BERT embeddings without fine-tuning on the specific task are significantly inferior to the fine-tuned ones.  
% \citet{reimers2019sentence} shows that directly employing BERT to represent sentences does not work well in terms of semantic textual similarity. 
% Either averaged BERT context embeddings or the BERT [CLS] vector\footnote{Roughly the [CLS] vector is a weighted average of BERT context embeddings learned via next sentence prediction~\citep{devlin2018bert}} fails to beat averaged GloVe~\cite{pennington2014glove} embeddings, in the setting without fine-tuning.
% Note that Glove is a simple baseline proposed several years ago, while BERT is pretrained on massive data with very large model capacity. 
% This indicates that the BERT semantic space may not be fully exploited for the semantic similarity purpose.
% Nevertheless, the deficiency of the pretrained BERT embedding space for semantic textual similarity has not been understood yet. 

Towards this end, we first study the connection between the BERT pretraining objective and the semantic similarity task. Our analysis reveals that the sentence embeddings of BERT should be able to intuitively reflect the semantic similarity between sentences, which contradicts with experimental observations. Inspired by~\citet{gao2019representation} who find that the language modeling performance can be limited by the learned anisotropic word embedding space where the word embeddings occupy a narrow cone, and \citet{ethayarajh2019contextual} who find that BERT word embeddings also suffer from anisotropy, we hypothesize that the sentence embeddings from BERT -- as average of context embeddings from last layers\footnote{In this paper, we compute average of context embeddings from last one or two layers as our sentence embeddings since they are consistently better than the [CLS] vector as shown in~\citep{reimers2019sentence}.} -- may suffer from similar issues. 
Through empirical probing over the embeddings, we further observe that the BERT sentence embedding space is semantically non-smoothing and poorly defined in some areas, which makes it hard to be used directly through simple similarity metrics such as dot product or cosine similarity. 

To address these issues, we propose to transform the BERT sentence embedding distribution into a smooth and isotropic Gaussian distribution through normalizing flows~\citep{dinh2014nice}, which is an invertible function parameterized by neural networks. 
%
% In this paper, we propose a novel and effective unsupervised calibration method to the BERT embedding space, which is motivated by previous studies of isotropic embedding space~\citep{arora2016simple,mu2017all,gao2019representation,wang2020spectrum,ethayarajh2019contextual}.
Concretely, we learn a flow-based generative model to maximize the likelihood of generating BERT sentence embeddings from a standard Gaussian latent variable in a \emph{unsupervised} fashion. During training, only the flow network is optimized while the BERT parameters remain unchanged. The learned flow, an invertible mapping function between the BERT sentence embedding and Gaussian latent variable, is then used to transform the BERT sentence embedding to the Gaussian space. We name the proposed method as \textit{BERT-flow}.
% Concretely, an invertible mapping to standard Gaussian is fitted over the marginal distribution of the BERT embedding space via maximum likelihood training. We instantiate the invertible mapping with normalizing flows. Thanks to the isotropy of standard Gaussian, the excessive influence of word frequency over the embedding space can get suppressed.

% To verify our hypothesis and the effectiveness of our proposed method, 
We perform extensive experiments on 7 standard semantic textual similarity benchmarks without using any downstream supervision. Our empirical results demonstrate that the flow transformation is able to consistently improve BERT by up to 12.70 points with an average of 8.16 points in terms of Spearman correlation between cosine embedding similarity and human annotated similarity. When combined with external supervision from natural language inference tasks~\citep{snli,mnli}, our method outperforms the sentence-BERT embeddings~\citep{reimers2019sentence}, leading to new state-of-the-art performance. In addition to semantic similarity tasks, we apply sentence embeddings to a question-answer entailment task, QNLI~\citep{wang2018glue}, directly without task-specific supervision, and demonstrate the superiority of our approach. 
% Specifically, our method can lead to 10+ Spearman correlation gain on STS-B for both siamese BERT-base and BERT-large. 
% Even for NLI-fine-tuned BERT models, which produce SOTA scores, our method can still further improve the performance by a large margin. 
% On QNLI, a question-answer entailment task adapted from SQUAD which comprises 100K+ sentence pairs, our method can also improve the AUC without any task-specific supervision compared to the BERT baselines.
Moreover, our further analysis implies that BERT-induced similarity can excessively correlate with lexical similarity compared to semantic similarity, and our proposed flow-based method can effectively remedy this problem.

\section{Understanding the Sentence Embedding Space of BERT}
\label{sec:2}
To encode a sentence into a fixed-length vector with BERT, it is a convention to either compute an average of context embeddings in the last few layers of BERT, or extract the BERT context embedding at the position of the [CLS] token. Note that there is no token masked when producing sentence embeddings, which is different from pretraining. 

\citet{reimers2019sentence} demonstrate that such BERT sentence embeddings lag behind the state-of-the-art sentence embeddings in terms of semantic similarity. On the STS-B dataset, BERT sentence embeddings are even less competitive to averaged GloVe~\citep{pennington2014glove} embeddings, which is a simple and non-contextualized baseline proposed several years ago. Nevertheless, this incompetence has not been well understood yet in existing literature. 

Note that as demonstrated by \citet{reimers2019sentence}, averaging context embeddings consistently outperforms the [CLS] embedding. Therefore, unless mentioned otherwise, we use average of context embeddings as BERT sentence embeddings and do not distinguish them in the rest of the paper.

\subsection{The Connection between Semantic Similarity and BERT Pre-training}
\label{sec:formalizing:semantic:similarity}
% \bl{TODO: Now that we are talking about two things: semantic similarity tasks and BERT training, but this entire 2.1 only talks about BERT. Better to briefly describe and formulate ``semantic similarity task'' ($h_c^Th_c^{\prime}$) in the beginning as well.}

We consider a sequence of tokens $x_{1:T} = (x_1, \ldots , x_T )$. Language modeling (LM) factorizes the joint probability $p(x_{1:T})$ in an autoregressive way, namely $\log p(x_{1:T}) = \sum_{t=1}^T \log p(x_t | c_t)$ where the context $c_t = x_{1:t-1}$. To capture bidirectional context during pretraining, BERT proposes a masked language modeling (MLM) objective, which instead factorizes the probability of noisy reconstruction $p(\bar{x} | \hat{x}) = \sum_{t=1}^T {m_t}~ p(x_t | c_t)$, where $\hat{x}$ is a corrupted sequence, $\bar{x}$ is the masked tokens, $m_t$ is equal to 1 when $x_t$ is masked and 0 otherwise. The context $c_t = \hat{x}$.

%\paragraph{Reducing LM/MLM to $p(x | c)$. }
Note that both LM and MLM can be reduced to modeling the conditional distribution of a token $x$ given the context $c$, which is typically formulated with a softmax function as,

\begin{equation}
    \label{eq:softmax}
    p(x | c) = \frac{\exp \rvh_c^\top \rvw_x}{\sum_{x'} \exp \rvh_c^\top \rvw_{x'}} .
\end{equation}

Here the context embedding $\rvh_c$ is a function of $c$, which is usually heavily parameterized by a deep neural network (e.g., a Transformer~\citep{vaswani2017attention}); The word embedding $\rvw_x$ is a function of $x$, which is parameterized by an embedding lookup table. 

The similarity between BERT sentence embeddings can be reduced to the similarity between BERT context embeddings $\rvh_c^T \rvh_{c'}$\footnote{This is because we approximate BERT sentence embeddings with context embeddings, and compute their dot product (or cosine similarity) as model-predicted sentence similarity. Dot product is equivalent to cosine similarity when the embeddings are normalized to unit hyper-sphere.}.
However, as shown in Equation~\ref{eq:softmax}, the pretraining of BERT does not explicitly involve the computation of $\rvh_c^T \rvh_{c'}$. Therefore, we can hardly derive a mathematical formulation of what $\rvh_c^\top \rvh_{c'}$ exactly represents.

\paragraph{Co-Occurrence Statistics as the Proxy for Semantic Similarity} 
Instead of directly analyzing $\rvh_c^T \rvh_c^{\prime}$, we consider $\rvh_c^\top \rvw_x$, the dot product between a context embedding $\rvh_c$ and a word embedding $\rvw_x$. According to \citet{yang2017breaking}, in a well-trained language model, $\rvh_c^\top \rvw_x$ can be approximately decomposed as follows,

\begin{align}
    \vspace{-10pt}
    \rvh_{c}^\top \rvw_{x} &\approx  \log p^* (x | c) + \lambda_{c}  \\
    &= \textrm{PMI}(x, c) + \log p(x) + \lambda_{c} . \label{eq:pmi}  
\end{align}
where $\textrm{PMI}(x, c) = \log \frac{p(x, c)}{p(x) p(c)}$ denotes the pointwise mutual information between $x$ and $c$, $\log p(x)$ is a word-specific term, and $\lambda_{c}$ is a context-specific term.

PMI captures how frequently two events co-occur more than if they independently occur. Note that co-occurrence statistics is a typical tool to deal with ``semantics'' in a computational way --- specifically, PMI is a common mathematical surrogate to approximate word-level semantic similarity~\cite{levy2014neural,ethayarajh2019towards}. Therefore, roughly speaking, it is semantically meaningful to compute the dot product between a context embedding and a word embedding.

\paragraph{Higher-Order Co-Occurrence Statistics as Context-Context Semantic Similarity.}
% \bl{TODO: this entire subsection is full of subjective arguments. try to make them more mathematically sound} 
During pretraining, the semantic relationship between two contexts $c$ and $c'$ could be inferred and reinforced with their connections to words. To be specific, if both the contexts $c$ and $c'$ co-occur with the same word $w$, the two contexts are likely to share similar semantic meaning. During the training dynamics, when $c$ and $w$ occur at the same time, the embeddings $h_c$ and $x_w$ are encouraged to be closer to each other, meanwhile the embedding $h_c$ and $x_{w'}$ where $w' \neq w$ are encouraged to be away from each other due to normalization. A similar scenario applies to the context $c'$. In this way, the similarity between $h_{c}$ and $h_{c'}$ is also promoted. With all the words in the vocabulary acting as hubs, the context embeddings should be aware of its semantic relatedness to each other. 

Higher-order context-context co-occurrence could also be inferred and propagated during pretraining. The update of a context embedding $h_c$ could affect another context embedding $h_{c'}$ in the above way, and similarly $h_{c'}$ can further affect another $h_{c''}$. Therefore, the context embeddings can form an implicit interaction among themselves via higher-order co-occurrence relations.

\begin{table*}[!h]
\centering
\scalebox{0.8}{
\begin{tabular}{lrrrrrr}
\toprule
Rank of word frequency   & $(0,100)$ & $[100,500)$ & $[500,5\textrm{K})$  & $[5\textrm{K},1\textrm{K})$  \\
\midrule\midrule
Mean $\ell_2$-norm & 0.95 & 1.04 & 1.22 & 1.45\\
\midrule\midrule
Mean $k$-NN $\ell_2$-dist. ($k=3$) & 0.77 & 0.93 & 1.16 & 1.30 \\
Mean $k$-NN $\ell_2$-dist. ($k=5$) & 0.83 & 0.99 & 1.22 & 1.34 \\
Mean $k$-NN $\ell_2$-dist. ($k=7$) & 0.87 & 1.04 & 1.26 & 1.37\\
\midrule
Mean $k$-NN dot-product. ($k=3$) & 0.73 & 0.92 & 1.20 & 1.63 \\
Mean $k$-NN dot-product. ($k=5$) & 0.73 & 0.91 & 1.19 & 1.61 \\
Mean $k$-NN dot-product. ($k=7$) & 0.72 & 0.90 & 1.17 & 1.60 \\
\bottomrule
\end{tabular}
}
\caption{\label{tab:wordfreq} The mean $\ell_2$-norm, as well as their distance to their $k$-nearest neighbors (among all the word embeddings) of the word embeddings of BERT, segmented by ranges of word frequency rank (counted based on Wikipedia dump; the smaller the more frequent).  }
%  The word frequency is counted based on a 1.8M-sentence subset randomly sampled from the Wikipedia dump.
%  of 04/04/2019; The way of dividing the ranges of word frequency is borrowed from \cite{mimno2017strange}.

\vspace{-10pt}
\end{table*}

\subsection{Anisotropic Embedding Space Induces Poor Semantic Similarity}
\label{sec:anisotropy}

As discussed in Section~\ref{sec:formalizing:semantic:similarity}, the pretraining of BERT should have encouraged semantically meaningful context embeddings implicitly. Why BERT sentence embeddings without finetuning yield unsatisfactory performance? 

% \jhc{In this section, we find out that (masked) language models tend to learn an anisotropic embedding space, and we hypothesize that the anisotropy should be responsible for the failure of learning a semantically meaningful embedding space}{directly say what we found, don't say we find/observe the anisotropic thing, refer to introduction for an example}.

To investigate the underlying problem of the failure, we use word embeddings as a surrogate because words and contexts share the same embedding space. If the word embeddings exhibits some misleading properties, the context embeddings will also be problematic, and vice versa.

% \jh{you can clarify you are using word embeddings as a surrogate and why? Because readers would think the ``embedding space'' is sentence embedding space but it is actually word embedding space. I can understand sentence embedding is just average of word emb, but needs one or two sentences to clarify} 

\citet{gao2019representation} and \citet{wang2020spectrum} have pointed out that, for language modeling, the maximum likelihood training with Equation~\ref{eq:softmax} usually produces an anisotropic word embedding space. ``Anisotropic'' means word embeddings occupy a narrow cone in the vector space.
% \jh{maybe briefly describe what a anisotropic cone is in the context of word embeddings?}. 
This phenomenon is also observed in the pretrained Transformers like BERT, GPT-2, etc~\cite{ethayarajh2019contextual}. 

In addition, we have two empirical observations over the learned anisotropic embedding space.

\paragraph{Observation 1: Word Frequency Biases the Embedding Space} 
\label{sec:h1}
% \jh{why "high-freq close to origin and low-freq far from origin" entails anisotropy? }

We expect the embedding-induced similarity to be consistent to semantic similarity. If embeddings are distributed in different regions according to frequency statistics, the induced similarity is not useful any more.

However, as discussed by \citet{gao2019representation}, anisotropy is highly relevant to the imbalance of word frequency. They prove that under some assumptions, the optimal embeddings of non-appeared tokens in Transformer language models can be extremely far away from the origin. They also try to roughly generalize this conclusion to rarely-appeared words. 
%\jh{any citation or intuition to make such claim?}. 

To verify this hypothesis in the context of BERT, we compute the mean $\ell_2$ distance between the BERT word embeddings and the origin (i.e., the mean $\ell_2$-norm). In the upper half of Table~\ref{tab:wordfreq}, we observe that high-frequency words are all close to the origin, while low-frequency words are far away from the origin.
% \jh{describe the experiment setting and motivation before talk about the results. For example, ``to verify this hypothesis, we compute the mean ....''}. 

This observation indicates that the word embeddings can be biased to word frequency. This coincides with the second term in Equation~\ref{eq:pmi}, the log density of words. 
Because word embeddings play a role of connecting the context embeddings during training, context embeddings might be misled by the word frequency information accordingly and its preserved semantic information can be corrupted.
% \jh{maybe mention this earlier in the section as well, you need motivations to say why are you analyzing word embedding space}.

\paragraph{Observation 2: Low-Frequency Words Disperse Sparsely} 
\label{sec:h2}
We observe that, in the learned anisotropic embedding space, high-frequency words concentrates densely and low-frequency words disperse sparsely. 

This observation is achieved by computing the mean $\ell_2$ distance of word embeddings to their $k$-nearest neighbors. In the lower half of Table~\ref{tab:wordfreq}, we observe that the embeddings of low-frequency words tends to be farther to their $k$-NN neighbors compared to the embeddings of high-frequency words. This demonstrates that low-frequency words tends to disperse sparsely. 

% \jh{I don't understand why holes in the embedding space are bad if all word embeddings are well-defined already? I can't follow the logic of this paragraph, maybe explain}
Due to the sparsity, many ``holes'' could be formed around the low-frequency word embeddings in the embedding space, where the semantic meaning can be poorly defined. Note that BERT sentence embeddings are produced by averaging the context embeddings, which is a convexity-preserving operation. However, the holes violate the convexity of the embedding space. This is a common problem in the context of representation learining~\citep{rezende2018taming,li2019surprisingly,ghosh2019variational}. Therefore, the resulted sentence embeddings can locate in the poorly-defined areas, and the induced  similarity can be problematic.

% However, note that low-frequency words play an important role for learning higher-order co-occurrence statistics. Imagine that two contexts $c_1$ and $c_2$ co-occur with a stop-word \textit{the}, we will not expect the contexts share similar semantic meaning; but if the two contexts co-occur with some common low-frequency word, for example, \textit{diabetes}, they are more likely to be semantically similar to each other. Therefore, the induced context-context semantic similarity can be problematic\jh{mention your hypothesis again here}.

\section{Proposed Method: BERT-flow}
To verify the hypotheses proposed in Section~\ref{sec:anisotropy}, and to circumvent the incompetence of the BERT sentence embeddings, we proposed a calibration method called BERT-flow in which we take advantage of \textit{an invertible mapping from the BERT embedding space to a standard Gaussian latent space}. The invertibility condition assures that the mutual information between the embedding space and the data examples does not change. 

\begin{figure}[!t]
	\centering
	\includegraphics[width=\columnwidth]{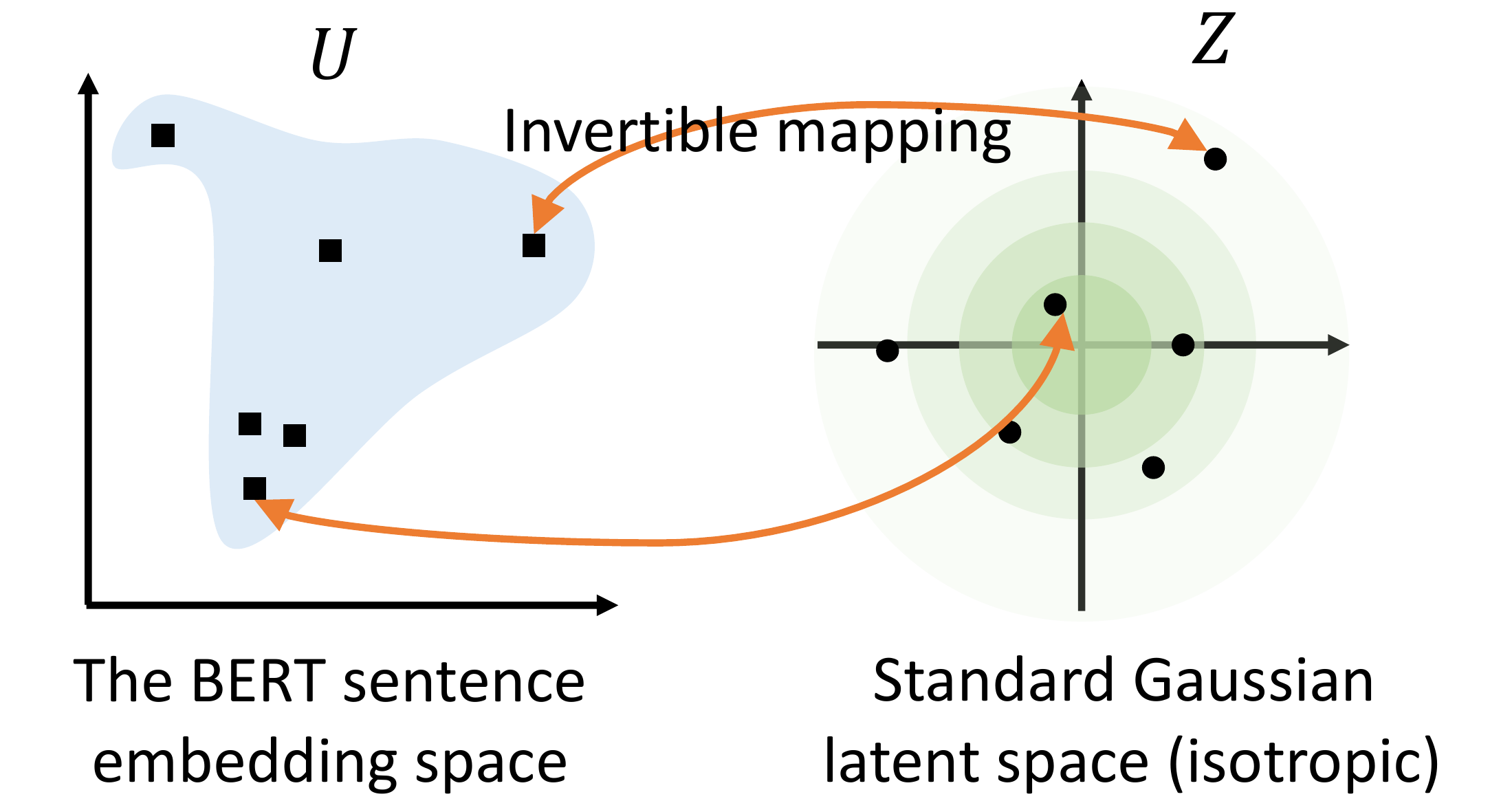}
	\vspace{-20pt}
	\caption{\label{fig:flow} An illustration of our proposed flow-based calibration over the original sentence embedding space of BERT. }
	\vspace{-10pt}
\end{figure}

\subsection{Motivation}
A standard Gaussian latent space may have favorable properties which can help with our problem. 

\paragraph{Connection to Observation 1} First, standard Gaussian satisfies isotropy. The probabilistic density in standard Gaussian distribution does not vary in terms of angle. If the $\ell_2$ norm of samples from standard Gaussian are normalized to 1, these samples can be regarded as uniformly distributed over a unit sphere.

We can also understand the isotropy from a singular spectrum perspective. As discussed above, the anisotropy of the embedding space stems from the imbalance of word frequency. In the literature of traditional word embeddings, \citet{mu2017all} discovers that the dominating singular vectors can be highly correlated to word frequency, which misleads the embedding space. By fitting a mapping to an isotropic distribution, the singular spectrum of the embedding space can be flattened. In this way, the word frequency-related singular directions, which are the dominating ones, can be suppressed. 

\paragraph{Connection to Observation 2} Second, the probabilistic density of Gaussian is well defined over the entire real space. This means there are no ``hole'' areas, which are poorly defined in terms of probability. The helpfulness of Gaussian prior for mitigating the ``hole'' problem has been widely observed in existing literature of deep latent variable models~\citep{rezende2018taming,li2019surprisingly,ghosh2019variational}.

% % The idea of flattening the singular spectrum \citet{wang2020spectrum} also develops the idea of . However, its method requires to be adopted during training, while training BERT again from scratch is hardly feasible. 

% \paragraph{Balancing the Margins (Section~\ref{sec:h2}).}
% We expect such transformation can balance the margin among the context embeddings. Hopefully, the context embeddings that are too far away from others can be contracted to the interior of the convex hull of all the context embedding, at the same time too-adjacent context embeddings are be pulled away to each other. In this way, it is possible that the strong lexical proximity can be adjusted to be moderate. \bl{not rigorous. need to be rewrite}

\subsection{Flow-based Generative Model}
% \bl{TODO: introduce more about flow}\jh{emphasize that the flow generative model learning is unsupervised in this section. Also need to mention it in introduction. And emphasize that only the flow transformation is trained using the objective not the BERT, this potentially makes the unsupervised training and adaptation very fast. }
% 

We instantiate the invertible mapping with flows. A flow-based generative model~\cite{kobyzev2019normalizing} establishes an invertible transformation from the latent space $\mathcal{Z}$ to the observed space $\mathcal{U}$. The generative story of the model is defined as
\begin{equation*}
    \rvz \sim p_{\mathcal{Z}}(\rvz), \rvu = f_{\phi}(\rvz)
    \vspace{-5pt}
\end{equation*}
where $\rvz \sim p_{\mathcal{Z}} (\mathbf{\rvz})$ the prior distribution, and $f: \mathcal{Z} \rightarrow \mathcal{U}$ is an invertible transformation. With the change-of-variables theorem, the probabilistic density function (PDF) of the observable $x$ is given as, 
\begin{equation*}
    \label{eq:flow}
    p_{\mathcal{U}} (\rvu) = p_{\mathcal{Z}} (f_\phi^{-1} (\rvu)) 
    ~\lvert \textrm{det} \frac{\partial f_\phi^{-1} (\rvu)}{\partial \rvu} \rvert
\end{equation*}

In our method, we learn a flow-based generative model
by maximizing the likelihood of generating BERT sentence embeddings from a standard Gaussian latent latent variable. In other words, the base distribution $p_{\mathcal{Z}}$ is a standard Gaussian and we consider the extracted BERT sentence embeddings as the observed space $\mathcal{U}$. We maximize the likelihood of $\mathcal{U}$'s marginal via Equation~\ref{eq:flow:mle} in a fully \emph{unsupervised} way.
\begin{align}
    \vspace{-10pt}
    \label{eq:flow:mle}
    \textrm{max}_{\phi}~
    & \mathbb{E}_{ \rvu = \textrm{BERT(sentence)}, \textrm{sentence} \sim \mathcal{D} }
      \nonumber \\
    & \log p_{\mathcal{Z}} (f_\phi^{-1} (\rvu)) + \log
    ~\lvert \textrm{det} \frac{\partial f_\phi^{-1} (\rvu)}{\partial \rvu} \rvert, 
    \vspace{-5pt}
\end{align}
Here $\mathcal{D}$ denotes the dataset, in other words, the collection of sentences. Note that during training, only the flow parameters are optimized while the BERT parameters remain unchanged. Eventually, we learn an invertible mapping function $f_\phi^{-1}$ which can transform each BERT sentence embedding $\rvu$ into a latent Gaussian representation $\rvz$ without loss of information.

The invertible mapping $f_\phi$ is parameterized as a neural network, and the architectures are usually carefully designed to guarantee the invertibility~\cite{dinh2014nice}. Moreover, its determinant $\lvert \textrm{det} \frac{\partial f_\phi^{-1} (\rvu)}{\partial \rvu} \rvert$  should also be easy to compute so as to make the maximum likelihood training tractable. In our experiments, we follows the design of Glow~\cite{kingma2018glow}. The Glow model is composed of a stack of multiple invertible transformations, namely \textit{actnorm}, \textit{invertible $1\times1$ convolution}, and \textit{affine coupling layer}\footnote{For concrete mathamatical formulations, please refer to Table 1 of \citet{kingma2018glow}}.
% \jh{briefly explain what is glow transformation here, maybe with equations. More details can go appendix}
We simplify the model by replacing affine coupling with additive coupling~\cite{dinh2014nice} to reduce model complexity, and replacing the invertible $1\times1$ convolution with random permutation to avoid numerical errors. For the mathematical formula of the flow model with additive coupling, please refer to Appendix~\ref{sec:appendix:additive:coupling}.

\section{Experiments}

\begin{table*}[!h]
\begin{center}
\scalebox{0.8}{
\begin{tabular}{l llllllll}
\toprule

% \thead{Dataset \\ \#pairs (test/full)} & \thead{STS-B \\ 1.4K / 8.6K } & \thead{SICK-R \\ 4.9K / 9.9K} & \thead{STS-12 \\ 3.1K / - } & \thead{STS-13 \\ 1.5K / -} & \thead{STS-14 \\ 3.7K / -} & \thead{STS-15 \\ 8.5K / - } & \thead{STS-16 \\ 9.2K / -} \\

Dataset & STS-B & SICK-R & STS-12 & STS-13 & STS-14 & STS-15 & STS-16 \\
\midrule\midrule
\multicolumn{8}{c}{\emph{Published in \cite{reimers2019sentence}}}  \\
Avg. GloVe embeddings & 58.02 &	53.76 &	55.14 &	70.66 &	59.73 &	68.25 &	63.66 \\
Avg. BERT embeddings & 46.35 &	58.40 &	38.78 &	57.98 &	57.98 &	63.15 &	61.06 \\
BERT CLS-vector & 16.50 &	42.63 &	20.16 &	30.01 &	20.09 &	36.88 &	38.03 \\
\midrule\midrule
\multicolumn{8}{c}{\emph{Our Implementation}}  \\
BERT$_{\text{base}}$ & 47.29	& 58.21	& 49.07	& 55.92	& 54.75	& 62.75	& 65.19 \\
BERT$_{\text{base}}$-last2avg & 59.04	& 63.75	& 57.84	& 61.95	& 62.48	& 70.95	& 69.81 \\
BERT$_{\text{base}}$-flow (NLI$^*$) & 58.56 ($\mathcolor{red}{\downarrow}$)	& \bf 65.44 ($\mathcolor{green}{\uparrow}$)	& 59.54	($\mathcolor{green}{\uparrow}$) & 64.69 ($\mathcolor{green}{\uparrow}$)	& 64.66 ($\mathcolor{green}{\uparrow}$) & 72.92	($\mathcolor{green}{\uparrow}$) & 71.84 ($\mathcolor{green}{\uparrow}$) \\
BERT$_{\text{base}}$-flow (target) & 70.72  ($\mathcolor{green}{\uparrow}$)	& 63.11($\mathcolor{red}{\downarrow}$) 	& 63.48 ($\mathcolor{green}{\uparrow}$)	& 72.14	 ($\mathcolor{green}{\uparrow}$) & 68.42	 ($\mathcolor{green}{\uparrow}$) & 73.77	 ($\mathcolor{green}{\uparrow}$) & 75.37  ($\mathcolor{green}{\uparrow}$) \\
% ~~ + flow-cat-STS-12-16 & -	    & -	& 63.02	 ($\mathcolor{green}{\uparrow}$) & 72.43	 ($\mathcolor{green}{\uparrow}$) & 68.80	 ($\mathcolor{green}{\uparrow}$) & 74.27	 ($\mathcolor{green}{\uparrow}$) & 75.28  ($\mathcolor{green}{\uparrow}$) \\
\midrule
BERT$_{\text{large}}$ & 46.99	& 53.74	& 46.89	& 53.32	& 49.27	& 56.54	& 61.63 \\
BERT$_{\text{large}}$-last2avg & 59.56	& 60.22	& 57.68	& 61.37	& 61.02	& 68.04	& 70.32 \\
BERT$_{\text{large}}$-flow (NLI$^*$) & 68.09 ($\mathcolor{green}{\uparrow}$)	& 64.62	($\mathcolor{green}{\uparrow}$) & 61.72 ($\mathcolor{green}{\uparrow}$)	& 66.05 ($\mathcolor{green}{\uparrow}$)	& 66.34	($\mathcolor{green}{\uparrow}$) & 74.87	($\mathcolor{green}{\uparrow}$) & 74.47 ($\mathcolor{green}{\uparrow}$) \\
BERT$_{\text{large}}$-flow (target) & \bf 72.26 ($\mathcolor{green}{\uparrow}$)	& 62.50	($\mathcolor{green}{\uparrow}$) & \bf 65.20	($\mathcolor{green}{\uparrow}$) & \bf 73.39 ($\mathcolor{green}{\uparrow}$)	& \bf 69.42 ($\mathcolor{green}{\uparrow}$)	& \bf 74.92 ($\mathcolor{green}{\uparrow}$)	& \bf 77.63 ($\mathcolor{green}{\uparrow}$) \\
% ~~ + flow-cat-STS-12-16 & -	    & -	& 63.78 ($\mathcolor{green}{\uparrow}$)	& \bf 73.69	($\mathcolor{green}{\uparrow}$) & \bf 69.81 ($\mathcolor{green}{\uparrow}$)	& \bf 75.45 ($\mathcolor{green}{\uparrow}$)	& 77.08 ($\mathcolor{green}{\uparrow}$)\\
\bottomrule
\end{tabular}
}
\caption{
Experimental results on semantic textual similarity \textbf{without} using NLI supervision.
We report the Spearman's rank correlation between the cosine similarity of sentence embeddings and the gold labels on multiple datasets. Numbers are reported as $\rho \times 100$. $\mathcolor{green}{\uparrow}$ denotes outperformance over its BERT baseline and $\mathcolor{red}{\downarrow}$ denotes underperformance. Our proposed BERT-flow method achieves the best scores. Note that our BERT-flow use \textit{-last2avg} as default setting. $*$: Use NLI corpus for the unsupervised training of flow; supervision labels of NLI are NOT visible.  }
\label{tbl:exp:no:nli}
\end{center}
\vspace{-20pt}
\end{table*}

To verify our hypotheses and demonstrate the effectiveness of our proposed method, in this section we present our experimental results for various tasks related to semantic textual similarity under multiple configurations. For the implementation details of our siamese BERT models and flow-based models, please refer to Appendix~\ref{sec:appendix:implementation}.

% \bl{TODO: roberta; classification}

\subsection{Semantic Textual Similarity}

\paragraph{Datasets.}
We evaluate our approach extensively on the semantic textual similarity (STS) tasks. We report results on 7 datasets, namely the STS benchmark (STS-B)~\citep{cer2017semeval}
the SICK-Relatedness (SICK-R) dataset~\citep{marelli2014sick} and the
STS tasks 2012 - 2016~\citep{agirre2012semeval,agirre2013sem,agirre2014semeval,agirre2015semeval,agirre2016semeval}. We obtain all these datasets via the SentEval toolkit~\citep{conneau2018senteval}. These datasets provide a fine-grained gold standard semantic similarity between 0 and 5 for each sentence pair.

% Among these datasets, STS-B is the newest one and is carefully downsampled from the corpus of STS 2012 - 2017. It has been selected in the GLUE benchmark~\citep{wang2018glue}, which is widely used for comparing the lastest language understanding approaches. 
% Sentence pairs of SICK-R are also annotated for textual entailment (SICK-E), hence the data might be collected to be biased to entailment. 

\paragraph{Evaluation Procedure. }
Following the procedure in previous work like Sentence-BERT~\cite{reimers2019sentence} for the STS task, the prediction of similarity consists of two steps: (1) first, we obtain sentence embeddings for each sentence with a sentence encoder, and  
% For BERT models, we consider both BERT$_{\text{base}}$ and BERT$_{\text{large}}$. In this work we use an average pooling over BERT context embeddings in the last one or two layers as the sentence embedding which is found to outperform the [CLS] vector as explained in the beginning of Section~\ref{sec:2}. Interestingly, in our preliminary exploration, we find out that averaging the last two layers of BERT (denoted by \textit{-last2avg}) consistently produce better results compared to only averaging the last one layer. Therefore, we choose \textit{-last2avg} as our default configuration when verifying our own approach.
(2) then, we compute the cosine similarity between the two embeddings of the input sentence pair as our model-predicted similarity. The reported numbers are the Spearman's correlation coefficients between the predicted similarity and gold standard similarity scores, which is the same way as in \citep{reimers2019sentence}.

\paragraph{Experimental Details. }
We consider both BERT$_{\text{base}}$ and BERT$_{\text{large}}$ in our experiments. Specifically, we use an average pooling over BERT context embeddings in the last one or two layers as the sentence embedding which is found to outperform the [CLS] vector. Interestingly, our preliminary exploration shows that averaging the last two layers of BERT (denoted by \textit{-last2avg}) consistently produce better results compared to only averaging the last one layer. Therefore, we choose \textit{-last2avg} as our default configuration when assessing our own approach.

For the proposed method, the flow-based objective (Equation~\ref{eq:flow:mle}) is maximized only to update the invertible mapping while the BERT parameters remains unchanged. Our flow models are by default learned over the full target dataset (train + validation + test).
% \jh{have you tried only using train? While this is unsupervised, it is still a bit unfair}. 
We denote this configuration as \textit{flow (target)}. Note that although we use the sentences of the entire target dataset, learning flow does not use any provided labels for training, thus it is a purely \emph{unsupervised} calibration over the BERT sentence embedding space.

We also test our flow-based model learned on a concatenation of SNLI~\citep{snli} and MNLI~\citep{mnli} for comparison (\emph{flow (NLI)}). The concatenated NLI datasets comprise of tremendously more sentence pairs (SNLI 570K + MNLI 433K). Note that ``flow (NLI)'' does not require any supervision label. When fitting flow on NLI corpora, we only use the raw sentences instead of the entailment labels. An intuition behind the flow (NLI) setting is that, compared to Wikipedia sentences (on which BERT is pretrained), the raw sentences of both NLI and STS are simpler and shorter. This means the NLI-STS discrepancy could be relatively smaller than the Wikipedia-STS discrepancy.

We run the experiments on two settings: (1) when external labeled data is unavailable. This is the natural setting where we learn flow parameters with the unsupervised objective (Equation~\ref{eq:flow:mle}), meanwhile BERT parameters are unchanged. (2) we first fine-tune BERT on the SNLI+MNLI textual entailment classification task in a siamese fashion~\citep{reimers2019sentence}. For BERT-flow, we further learn the flow parameters. This setting is to compare with the state-of-the-art results which utilize NLI supervision~\citep{reimers2019sentence}. We denote the two different models as BERT-NLI and BERT-NLI-flow respectively.

\begin{table*}[ht]
\begin{center}
\scalebox{0.8}{
\begin{tabular}{l llllllll}
\toprule

% \thead{Dataset \\ \#pairs (test/full)} & \thead{STS-B \\ 1.4K / 8.6K } & \thead{SICK-R \\ 4.9K / 9.9K} & \thead{STS-12 \\ 3.1K / - } & \thead{STS-13 \\ 1.5K / -} & \thead{STS-14 \\ 3.7K / -} & \thead{STS-15 \\ 8.5K / - } & \thead{STS-16 \\ 9.2K / -} \\

Dataset & STS-B & SICK-R & STS-12 & STS-13 & STS-14 & STS-15 & STS-16 \\
\midrule\midrule
\multicolumn{8}{c}{\emph{Published in \cite{reimers2019sentence}}}  \\

InferSent - Glove & 68.03 &	65.65 &	52.86 &	66.75 &	62.15 &	72.77 &	66.86 \\
USE & 74.92 &	76.69 &	64.49 &	67.80 &	64.61 &	76.83 &	73.18 \\
SBERT$_{\text{base}}$-NLI & 77.03 &	72.91 &	70.97 &	76.53 &	73.19 &	79.09 &	74.30 \\
SBERT$_{\text{large}}$-NLI & 79.23 &	73.75 &	72.27 &	78.46 &	74.90 &	80.99 &	76.25 \\
SRoBERTa$_{\text{base}}$-NLI & 77.77 &	74.46 &	71.54 &	72.49 &	70.80 &	78.74 &	73.69 \\
SRoBERTa$_{\text{large}}$-NLI & 79.10 &	74.29 &	\bf 74.53 &	77.00 &	73.18 &	81.85 &	76.82 \\

\midrule\midrule
\multicolumn{8}{c}{\emph{Our Implementation}}  \\
BERT$_{\text{base}}$-NLI  & 77.08	& 72.62	& 66.23	& 70.22	& 72.15	& 77.35	& 73.91 \\
BERT$_{\text{base}}$-NLI-last2avg & 78.03	& 74.07	& 68.37	& 72.44	& 73.98	& 79.15	& 75.39 \\
BERT$_{\text{base}}$-NLI-flow (NLI$^*$)   & 79.10 ($\mathcolor{green}{\uparrow}$)	& \bf 78.03 ($\mathcolor{green}{\uparrow}$)	& 67.75 ($\mathcolor{red}{\downarrow}$)	& 76.73 ($\mathcolor{green}{\uparrow}$)	& 75.53 ($\mathcolor{green}{\uparrow}$)	& 80.63 ($\mathcolor{green}{\uparrow}$)	& 77.58 ($\mathcolor{green}{\uparrow}$) \\
BERT$_{\text{base}}$-NLI-flow (target) & 81.03 ($\mathcolor{green}{\uparrow}$)	& 74.97 ($\mathcolor{green}{\uparrow}$)	& 68.95 ($\mathcolor{green}{\uparrow}$)	& 78.48 ($\mathcolor{green}{\uparrow}$)	& 77.62 ($\mathcolor{green}{\uparrow}$)	& 81.95 ($\mathcolor{green}{\uparrow}$)	& 78.94 ($\mathcolor{green}{\uparrow}$) \\
% ~~ + flow-cat-STS-12-16 & -	& -	& 69.74 ($\mathcolor{green}{\uparrow}$)	& 79.58 ($\mathcolor{green}{\uparrow}$)	& 77.78 ($\mathcolor{green}{\uparrow}$)	& 82.20 ($\mathcolor{green}{\uparrow}$)	& 80.04 ($\mathcolor{green}{\uparrow}$) \\
\midrule
BERT$_{\text{large}}$-NLI & 77.80	& 73.44	& 66.87	& 73.91	& 74.04	& 79.14	& 75.35 \\
BERT$_{\text{large}}$-NLI-last2avg & 78.45	& 74.93	& 68.69	& 75.63	& 75.55	& 80.35	& 76.81 \\
BERT$_{\text{large}}$-NLI-flow (NLI$^*$) & 79.89  ($\mathcolor{green}{\uparrow}$)	& 77.73  ($\mathcolor{green}{\uparrow}$)	& 69.61  ($\mathcolor{green}{\uparrow}$)	& 79.45  ($\mathcolor{green}{\uparrow}$)	& 77.56  ($\mathcolor{green}{\uparrow}$)	& 82.48	  ($\mathcolor{green}{\uparrow}$) & 79.36  ($\mathcolor{green}{\uparrow}$) \\
BERT$_{\text{large}}$-NLI-flow (target) & \bf 81.18  ($\mathcolor{green}{\uparrow}$)	& 74.52 ($\mathcolor{red}{\downarrow}$)	&  70.19	 ($\mathcolor{green}{\uparrow}$) & \bf 80.27  ($\mathcolor{green}{\uparrow}$)	& \bf 78.85  ($\mathcolor{green}{\uparrow}$)	& \bf 82.97  ($\mathcolor{green}{\uparrow}$)	& \bf 80.57  ($\mathcolor{green}{\uparrow}$)\\
% ~~ + flow-cat-STS-12-16 & -	& -	& 70.98  ($\mathcolor{green}{\uparrow}$)	& \bf  81.31  ($\mathcolor{green}{\uparrow}$) & \bf  79.10  ($\mathcolor{green}{\uparrow}$)	& \bf  83.33  ($\mathcolor{green}{\uparrow}$)	& \bf  81.63  ($\mathcolor{green}{\uparrow}$) \\

\bottomrule
\end{tabular}
}
\caption{
% \jh{Can we replace all the BERT-NLI-xxx with SBERT-NLL-xxx? Because they are essentially reimplementation of SBERT right? Also, Table 4 and Table 5 use SBERT + flow and we should be consistent}
Experimental results on semantic textual similarity \textit{with} NLI supervision. Note that our flows are still learned in a \textit{unsupervised} way. 
InferSent~\citep{conneau-EtAl:2017:EMNLP2017} is a siamese LSTM train on NLI, Universal Sentence Encoder (USE)~\citep{cer2018universal} replace the LSTM with a Transformer and SBERT~\cite{reimers2019sentence} further use BERT.
We report the Spearman's rank correlation between the cosine similarity of sentence embeddings and the gold labels on multiple datasets. Numbers are reported as $\rho \times 100$. $\mathcolor{green}{\uparrow}$ denotes outperformance over its BERT baseline and $\mathcolor{red}{\downarrow}$ denotes underperformance. Our proposed BERT-flow (i.e., the ``BERT-NLI-flow'' in this table) method achieves the best scores. Note that our BERT-flow use \textit{-last2avg} as default setting. $*$: Use NLI corpus for the unsupervised training of flow; supervision labels of NLI are NOT visible.}
\label{tbl:exp:nli}
\end{center}
\vspace{-10pt}
\end{table*}

\paragraph{Results w/o NLI Supervision. } 
As shown in Table~\ref{tbl:exp:no:nli}, the original BERT sentence embeddings (with both BERT$_{\text{base}}$ and BERT$_{\text{large}}$) fail to outperform the averaged GloVe embeddings. 
And averaging the last-two layers of the BERT model can consistently improve the results. For BERT$_{\text{base}}$ and BERT$_{\text{large}}$, our proposed flow-based method (BERT-flow (target)) can further boost the performance by 5.88 and 8.16 points on average respectively.
% base: 63.6885714286 --> 69.5728571429
% large: 62.6014285714 --> 70.76
% 2.93 or 2.59 for NLI-supervised setting
%
For most of the datasets, learning flows on the target datasets leads to larger performance improvement than on NLI. The only exception is SICK-R where training flows on NLI is better. We think this is because SICK-R is collected for both entailment and relatedness. 
Since SNLI and MNLI are also collected for textual entailment evaluation, the distribution discrepancy between SICK-R and NLI may be relatively small. Also due to the much larger size of the NLI datasets, it is not surprising that learning flows on NLI results in stronger performance.

\paragraph{Results w/ NLI Supervision.}
% Moreover, to compete with the state-of-the-art results, we conduct another round of experiments where the BERT models are first finetuned on SNLI+MNLI textual entailment classification in a siamese fashion~\cite{reimers2019sentence}.
Table~\ref{tbl:exp:nli} shows the results with NLI supervisions. Similar to the fully unsupervised results before, our isotropic embedding space from invertible transformation is able to consistently improve the SBERT baselines in most cases, and outperforms the state-of-the-art SBERT/SRoBERTa results by a large margin. Robustness analysis with respect to random seeds are provided in Appendix~\ref{sec:appendix:results:with:different:seeds}.
% on 6 out of 7 datasets, often

% Our implemented sentence-BERT performance on smaller datasets (STS-12-16) may not be as strong as their published numbers. The authors mentioned in \url{https://github.com/UKPLab/sentence-transformers/issues/50} that this is a common phenonmenon and might be related the random seed.

\vspace{-5pt}
\subsection{Unsupervised Question-Answer Entailment}
\vspace{-5pt}
% \jh{Is this experiment setting to eval sentence similarity first used in this paper? Any citations if not?}
In addition to the semantic textual similarity tasks, we  examine the effectiveness of our method on unsupervised question-answer entailment. We use Question Natural Language Inference (QNLI,~\citet{wang2018glue}), a dataset comprising 110K question-answer pairs (with 5K+ for testing). QNLI extracts the questions as well as their corresponding context sentences from SQUAD~\citep{rajpurkar2016squad}, and annotates each pair as either \textit{entailment} or \textit{no entailment}. 
In this paper, we further adapt QNLI as an unsupervised task. The similarity between a question and an answer can be predicted by computing the cosine similarity of their sentence embeddings. Then we regard \textit{entailment} as 1 and \textit{no entailment} as 0, and evaluate the performance of the methods with AUC.

% \jhc{}{To predict entailment through sentence embeddings directly without supervised fine-tuning, we compute the cosine similarity between the question and answer embeddings, and output entailment if the score > 0.5 and no entailment otherwise (not sure if this is what you are doing). Specifically, we take the SBERT with NLI supervision as our base encoder and fine-tune the flow transformation on the QNLI dataset without parallel information.} \jh{The experiment evaluation is unclear? How is similarity related to entailment decision? How to make the decision? Are you treating cosin sim > 0.5 as entailment and no entailment otherwise? If so, should explain.}\jhc{\jh{What is this metric? Is this some accuracy? Any citation? If it is accuracy, maybe write the full name}. 
% %Note that AUC is equivalent to the probability that entailment sentence pairs are ranked higher than no-entailment sentence pairs. 

As shown in Table~\ref{tbl:qnli}, our method consistently improves the AUC on the validation set of QNLI. Also, learning flow on the target dataset can produce superior results compared to learning flows on NLI.

% \begin{table}[!htb]
% \vspace{-5pt}
% \begin{minipage}{0.5\linewidth}
% \caption{\label{tbl:qnli} AUC on QNLI evaluated on the validation set.}
% \vspace{-7pt}
% \centering
% \scalebox{0.6}{
% \begin{tabular}{ll}
% \toprule
% \thead{Method} & \thead{AUC} \\ 
% \midrule
% SBERT-NLI-base-last2avg & 70.30\\
% ~~ + flow-NLI  & 72.52 ($\mathcolor{green}{\uparrow}$)  \\
% ~~ + flow-target & 76.17 ($\mathcolor{green}{\uparrow}$)  \\
% \midrule
% SBERT-NLI-large-last2avg & 70.41\\
% ~~ + flow-NLI  & 74.19 ($\mathcolor{green}{\uparrow}$)  \\
% ~~ + flow-target & \bf 77.09 ($\mathcolor{green}{\uparrow}$)  \\

% \bottomrule
% \end{tabular}}
% \end{minipage}%
% \begin{minipage}{0.5\linewidth}

% \caption{\label{tbl:baselines}Comparing Flow with baselines on STS-B. $k$ is selected among $1 \sim 20$ on the validation set. We report the Spearman’s rank correlation ($\times 100$).}

% \vspace{-7pt}
% \centering

% \scalebox{0.6}{
% \begin{tabular}{ll}
% \toprule
% Method & Correlation \\ 
% \midrule
% SBERT$_{\text{base}}$ & 47.29 \\
% ~~ + SN &  55.46  \\
% ~~ + NATSV ($k = 1$) & 51.79  \\
% ~~ + NATSV ($k = 10$) & 60.40  \\
% ~~ + SN + NATSV ($k = 1$) & 56.02  \\
% ~~ + SN + NATSV ($k = 6$) & 63.51  \\
% ~~ + flow-target  & \bf 65.62  \\

% \bottomrule
% \end{tabular}
% }
% \end{minipage} 
% \vspace{-10pt}
% \end{table}

\begin{table}[!h]
\centering
\scalebox{0.8}{
\begin{tabular}{ll}
\toprule
\thead{Method} & \thead{AUC} \\ 
\midrule
BERT$_{\text{base}}$-NLI-last2avg & 70.30\\
BERT$_{\text{base}}$-NLI-flow (NLI$^*$) & 72.52 ($\mathcolor{green}{\uparrow}$)  \\
BERT$_{\text{base}}$-NLI-flow (target) & 76.17 ($\mathcolor{green}{\uparrow}$)  \\
\midrule
BERT$_{\text{large}}$-NLI-last2avg & 70.41\\
BERT$_{\text{large}}$-NLI-flow (NLI$^*$)  & 74.19 ($\mathcolor{green}{\uparrow}$)  \\
BERT$_{\text{large}}$-NLI-flow (target) & \bf 77.09 ($\mathcolor{green}{\uparrow}$)  \\

\bottomrule
\end{tabular}
}
\caption{AUC on QNLI evaluated on the validation set. $*$: Use NLI corpus for the unsupervised training of flow; supervision labels of NLI are NOT visible.}
\label{tbl:qnli}
\vspace{-10pt}
\end{table}

\subsection{Comparison with Other Embedding Calibration Baselines}
\begin{table}[!h]
\centering
\scalebox{0.8}{
\begin{tabular}{ll}
\toprule
Method & Correlation \\ 
\midrule
BERT$_{\text{base}}$ & 47.29 \\
~~ + SN &  55.46  \\
~~ + NATSV ($k = 1$) & 51.79  \\
~~ + NATSV ($k = 10$) & 60.40  \\
~~ + SN + NATSV ($k = 1$) & 56.02  \\
~~ + SN + NATSV ($k = 6$) & 63.51  \\
BERT$_{\text{base}}$-flow (target)  & \bf 65.62  \\

\bottomrule
\end{tabular}
}
\caption{Comparing flow-based method with baselines on STS-B. $k$ is selected among $1 \sim 20$ on the validation set. We report the Spearman’s rank correlation ($\times 100$).}
\label{tbl:baselines}
\vspace{-10pt}
\end{table}

In the literature of traditional word embeddings, \citet{arora2016simple} and~\citet{mu2017all} also discover the anisotropy phenomenon of the embedding space, and they provide several methods to encourage isotropy:

\paragraph{Standard Normalization (SN).}
In this idea, we conduct a simple post-processing over the embeddings by computing the mean $\vect{\mu}$ and standard deviation $\vect{\sigma}$ of the sentence embeddings $\rvu$'s, and normalizing the embeddings by $\frac{\rvu - \vect{\mu}}{\vect{\sigma}}$.

\paragraph{Nulling Away Top-$k$ Singular Vectors (NATSV).} 
\citet{mu2017all} find out that sentence embeddings computed by averaging traditional word embeddings  tend to have a fast-decaying singular spectrum. They claim that, by nulling away the top-$k$ singular vectors, the anisotropy of the embeddings can be circumvented and better semantic similarity performance can be achieved.

We compare with these embedding calibration methods on STS-B dataset and the results are shown in Table~\ref{tbl:baselines}.
Standard normalization (SN) helps improve the performance but it falls behind nulling away top-$k$ singular vectors (NATSV). This means standard normalization cannot fundamentally eliminate the anisotropy. By combining the two methods, and carefully tuning $k$ over the validation set, further improvements can be achieved. Nevertheless, our method still produces much better results. We argue that NATSV can help eliminate anisotropy but it may also discard some useful information contained in the nulled vectors. On the contrary, our method directly learns an invertible mapping to isotropic latent space without discarding any information. 
% Therefore, we suggest choosing our proposed flow-based method to calibrate the BERT sentence embeddings.   

%  \jh{this figure can be improved if possible, for example, as three subfigures}
\begin{figure*}[!t]
	\centering
	\includegraphics[width=0.6\textwidth]{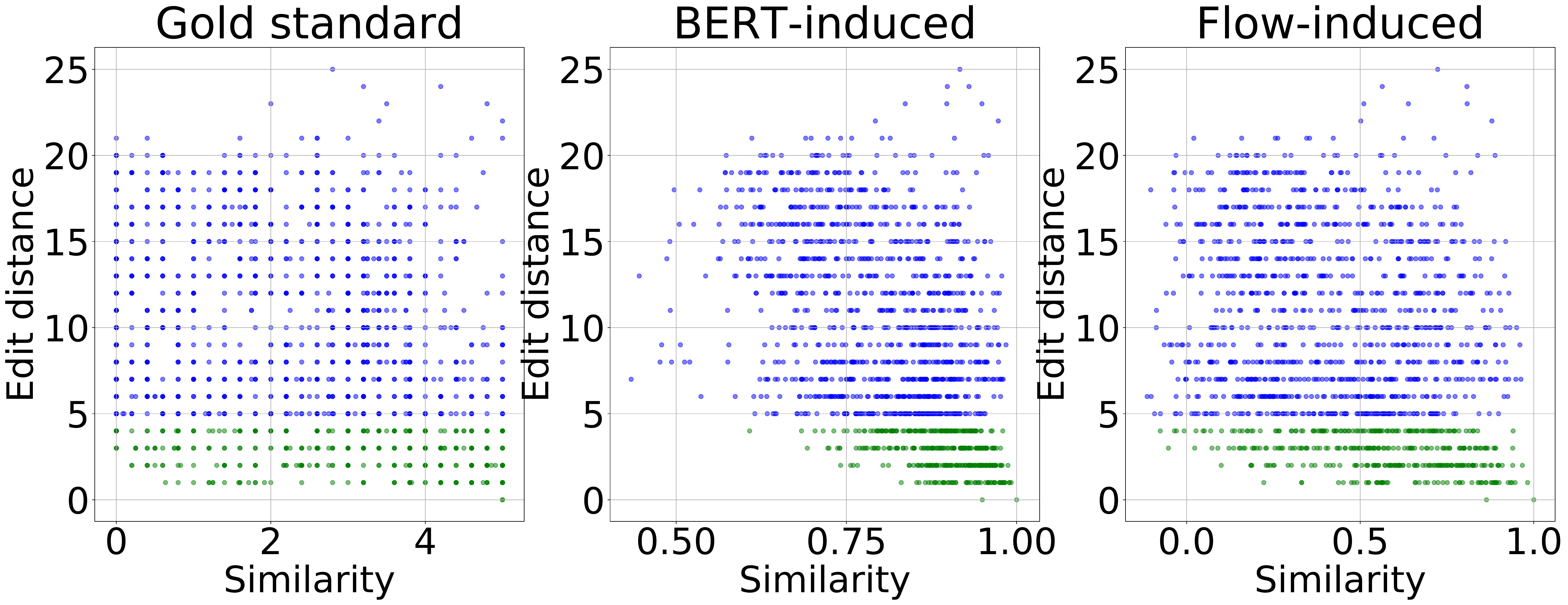}
	\caption{\label{fig:sim_ed} A scatterplot of sentence pairs, where the horizontal axis represents similarity (either gold standard semantic similarity or embedding-induced similarity), the vertical axis represents edit distance. The sentence pairs with edit distance $\leq 4$ are highlighted with {\color{green} green}, meanwhile the rest of the pairs are colored with {\color{blue} blue}. We can observed that lexically similar sentence pairs tends to be predicted to be similar by BERT embeddings, especially for the green pairs. Such correlation is less evident for gold standard labels or flow-induced embeddings. }
	\vspace{-10pt}
\end{figure*}

\subsection{Dicussion: Semantic Similarity Versus Lexical Similarity }
\label{sec:lexical:similarity}

% By ``lexical proximity'', we mean that, for sentences with smaller lexical difference, their embeddings are also closer to each other.

% \jhc{In addition to semantic similarity, lexical similarity may also need to be considered. It is possible to observe the correlation between BERT embeddings and the lexical space. First, BERT takes word embeddings as input and has skip connections in every Transformer layer. If we assume the word embeddings are lexically discriminative and the residual terms are not too large, we can roughly infer that the resulting BERT context embeddings are also lexically discriminative. Moreover, the random masking scheme of the MLM training explicitly assumes context embeddings should not be too sensitive to minor lexical perturbations.}{ }
In addition to semantic similarity, we further study lexical similarity induced by different sentence embeddings. Specifically, we use edit distance as the metric for lexical similarity between a pair of sentences, and focus on the correlations between the sentence similarity and edit distance. Concretely, we compute the cosine similarity in terms of BERT sentence embeddings as well as edit distance for each sentence pair. Within a dataset consisting of many sentence pairs, we compute the Spearman's correlation coefficient $\rho$ between the similarities and the edit distances, as well as between similarities from different models. We perform experiment on the STS-B dataset and include the human annotated gold similarity into this analysis.

\paragraph{BERT-Induced Similarity Excessively Correlates with Lexical Similarity.}
% We find out that BERT-induced similarity is strongly correlated to edit distance. 

% We focus on the correlation between BERT-induced similarity and edit distance. \jhc{First, for each sentence, we feed it into a BERT model and get its sentence embedding. Then for each sentence pair, we compute the cosine similarity between their embeddings as well as their edit distance.}{Concretely, we compute the cosine similarity in terms of BERT sentence embeddings as well as edit distance for each sentence pair.} Within a dataset consisting of many sentence pairs, we compute the Spearman's correlation coefficient $\rho$ between the similarities and the edit distances. 

% The second row in Table~\ref{tab:editdistance} demonstrates a highly negative correlation between BERT-induced similarity and edit distance. In other words, sentences which are close to each other in the lexical space are prone to also be close in the BERT embedding space. 
% The phenomenon of lexical proximity is especially evident for sentences with small lexical difference. In the middle plot of Figure~\ref{fig:sim_ed}, we find out that sentence pairs with edit distance $< 5$ tend to have a higher correlation between BERT-induced similarity and edit distance. 

\begin{table}[!t]
\centering
\scalebox{0.85}{
\begin{tabular}{lrr}
\toprule
Similarity & Edit distance &  Gold similarity \\
\midrule
Gold similarity & -24.61 & 100.00 \\
BERT-induce similarity & -50.49 & 59.30\\
Flow-induce similarity & -28.01 & 74.09 \\
\bottomrule
\end{tabular}
}
\caption{\label{tab:editdistance} Spearman's correlation $\rho$ between various sentence similarities on the validation set of STS-B. We can observe that BERT-induced similarity is highly correlated to edit distance, while the correlation with edit distance is less evident for gold standard or flow-induced similarity. }
\end{table}

% Generally, such lexical proximity might be favorable. It could provide useful hints for the BERT embedding space to acquire syntactic and semantic information. In addition to BERT, some other sentence representation learning methods~\cite{zhao2017learning,shen2019educating} assumes a good representation space should be insensitive to small lexical difference.

Table~\ref{tab:editdistance} shows that the correlation between BERT-induced similarity and edit distance is very strong ($\rho = -50.49$), considering that gold standard labels maintain a much smaller correlation with edit distance (\textbf{$\rho = -24.61$}). This phenomenon can also be observed in Figure~\ref{fig:sim_ed}. 
Especially, for sentence pairs with edit distance $\leq 4$ (highlighted with {\color{green} green}), BERT-induced similarity is extremely correlated to edit distance. However, it is not evident that gold standard semantic similarity correlates with edit distance. In other words, it is often the case where the semantics of a sentence can be dramatically changed by modifying a single word. For example, the sentences ``I like this restaurant'' and ``I dislike this restaurant'' only differ by one word, but convey opposite semantic meaning. BERT embeddings may fail in such cases. Therefore, we argue that the lexical proximity of BERT sentence embeddings is excessive, and can spoil their induced semantic similarity. 

\paragraph{Flow-Induced Similarity Exhibits Lower Correlation with Lexical Similarity.}
By transforming the original BERT sentence embeddings into the learned isotropic latent space with flow, the embedding-induced similarity not only aligned better with the gold semantic semantic similarity, but also shows a lower correlation with lexical similarity, as presented in the last row of Table~\ref{tab:editdistance}. The phenomenon is especially evident for the examples with edit distance $\leq 4$ (highlighted with {\color{green} green} in Figure~\ref{fig:sim_ed}).
%\jh{refer to Figure2? But I don't see it is especially evident with edit distance $\leq 4$, correlation seems to be lower for blue points}. 
This demonstrates that our proposed flow-based method can effectively suppress the excessive influence of lexical similarity over the embedding space.

\section{Conclusion and Future Work}
In this paper, we investigate the deficiency of the BERT sentence embeddings on semantic textual similarity, and propose a flow-based calibration which can effectively improve the performance. In the future, we are looking forward to diving in representation learning with flow-based generative models from a broader perspective.

\section*{Acknowledgments}
The authors would like to thank Jiangtao Feng, Wenxian Shi, Yuxuan Song, and anonymous reviewers for their helpful comments and suggestion on this paper.

\newpage
\bibliographystyle{acl_natbib}
\bibliography{references}

\newpage
\appendix
\onecolumn
\section{Mathematical Formula of the Invertible Mapping}
\label{sec:appendix:additive:coupling}
Generally, flow-based model is a stacked sequence of many invertible transformation layers: $f = f_1 \circ f_2 \circ \ldots \circ f_K$. Specifically, in our approach, each transformation $f_i: x \rightarrow y$ is an additive coupling layer, which can be mathematically formulated as follows.

\begin{align}
    \label{eq:additive:coupling:layer}
    y_{1:d} &= x_{1:d} \\
    y_{d+1:D} &= x_{d+1:D} + g_{\psi}(x_{1:d}). 
\end{align}

Here $g_{\psi}$ can be parameterized with a deep neural network for the sake of expressiveness. 

Its inverse function $f_i^{-1}: y \rightarrow x$ can be explicitly written as:
\begin{align}
    \label{eq:additive:coupling:layer:inverse}
    x_{1:d} &= y_{1:d} \\
    x_{d+1:D} &= y_{d+1:D} - g_{\psi}(y_{1:d}) .
\end{align}

\section{Implementation Details}
\label{sec:appendix:implementation}

Throughout our experiment, we adopt the official Tensorflow code of BERT~\footnote{\url{https://github.com/google-research/bert}} as our codebase. Note that we clip the maximum sequence length to 64 to reduce the costing of GPU memory. For the NLI finetuning of siamese BERT, we folllow the settings in~\citep{reimers2019sentence} (epochs = 1,  learning rate = $2e-5$, and batch size = 16). Our results may vary from their published one. The authors mentioned in \url{https://github.com/UKPLab/sentence-transformers/issues/50} that this is a common phenonmenon and might be related the random seed. Note that their implementation relies on the Transformers repository of Huggingface\footnote{\url{https://github.com/huggingface/transformers}}. This may also lead to discrepancy between the specific numbers.

% --max_seq_length=64 

% "--exp_name_prefix=exp_nonli \
%  --cached_dir=${CACHED_DIR} \
%  --sentence_embedding_type=avg-last-2 \
%  --flow=1 --flow_loss=1 \
%  --num_examples=0 \
%  --num_train_epochs=0.5 \
%  --flow_learning_rate=1e-3 \
%  --flow_model_config=${FLOW_CONFIG} \
%  --use_full_for_training=1"

Our implementation of flows is adapted from both the official repository of GLOW\footnote{\url{https://github.com/openai/glow}} as well as the implementation fo the Tensor2tensor library\footnote{\url{https://github.com/tensorflow/tensor2tensor/blob/master/tensor2tensor/models/research/glow.py}}. The hyperparameters of our flow models are given in Table~\ref{tab:flow:hyperparameter}. On the target datasets, we learn the flow parameters for 1 epoch with learning rate $1e-3$. On NLI datasets, we learn the flow parameters for 0.15 epoch with learning rate $2e-5$. The optimizer that we use is Adam. 

In our preliminary experiments on STS-B, we tune the hyperparameters on the dev set of STS-B. Empirically, the performance does not vary much with regard to the architectural hyperparameters compared to the learning schedule. Afterwards, we do not tune the hyperparameters any more when working on the other datasets. Empirically, we find the hyperparameters of flow are not sensitive across the datasets.

\begin{table}[!h]
\centering
% \scalebox{0.85}{
\begin{tabular}{ll}
\toprule
\bf Coupling architecture in  &  3-layer CNN with residual connection \\
\bf Coupling width & 32 \\
\bf \#levels & 2 \\
\bf Depth & 3 \\
\bottomrule
\end{tabular}
\caption{\label{tab:flow:hyperparameter} Flow hyperparameters.}
% }
\end{table}

% \section{Why Fit Flow on NLI Corpora?}
% \label{sec:appendix:why:fit:flow:on:NLI:corpora}
% Why fit flow on NLI corpora? Yes, NLI and STS are fundamentally different tasks. We conduct experiments with the NLI datasets because existing literature of sentence embeddings (including InferSent [1], Universal Sentence Encoder [2] and Sentence-BERT [3]) also employ NLI dataset when reporting the results on STS.
 
% Note that flow-NLI does not require any supervision label. When fitting flow on NLI corpora, we only use the raw sentences instead of the entailment labels. An intuition behind the flow-NLI setting is that, compared to Wikipedia sentences (on which BERT is pretrained), the raw sentences of both NLI and STS are simpler and shorter. This means the NLI-STS discrepancy could be relatively smaller than the Wikipedia-STS discrepancy.

\newpage
\section{Results with Different Random Seeds}
\label{sec:appendix:results:with:different:seeds}
We perform 5 runs with different random seeds in the NLI-supervised setting on STS-B. Results with standard deviation and median are demonstrated in Table~\ref{tbl:results:with:different:seeds}. Although the variance of NLI finetuning is not negligible, our proposed flow-based method consistently leads to improvement.

\begin{table}[!h]
\centering
\scalebox{1.0}{
\begin{tabular}{ll}
\toprule
\bf Method & \bf Spearman's $\rho$ \\ 
\midrule
BERT-NLI-large & 77.26 $\pm$ 1.76 (median: 78.19) \\
BERT-NLI-large-last2avg & 78.07 $\pm$ 1.50 (median: 78.68) \\
BERT-NLI-large-last2avg + flow-target & 81.10 $\pm$ 0.55 (median: 81.35) \\
\bottomrule
\end{tabular}
}
\caption{Results with different random seeds.}
\label{tbl:results:with:different:seeds}
\end{table}

\end{document}